\pdfoutput=1
\documentclass{article}

\PassOptionsToPackage{numbers, compress}{natbib}
\usepackage[final]{neurips_2022}
\usepackage{bbm}
\usepackage{graphicx}
\usepackage{amsmath}
\usepackage{amssymb}
\usepackage{booktabs}
\usepackage{listings}
\usepackage{subcaption}
\usepackage{algorithm}
\usepackage{textcomp}
\usepackage{xcolor}
\usepackage{xspace}
\usepackage{pdfrender}
\usepackage{algpseudocode}
\usepackage{enumitem}
\usepackage{multirow}

\usepackage[utf8]{inputenc} % allow utf-8 input
\usepackage[T1]{fontenc}    % use 8-bit T1 fonts
\usepackage{hyperref}       % hyperlinks
\usepackage{url}            % simple URL typesetting
\usepackage{booktabs}       % professional-quality tables
\usepackage{amsfonts}       % blackboard math symbols
\usepackage{nicefrac}       % compact symbols for 1/2, etc.
\usepackage{microtype}      % microtypography
\usepackage{xcolor}         % colors
\title{MapQA: A Dataset for Question Answering on Choropleth Maps}

\author{%
  Shuaichen Chang\thanks{The first two authors contributed equally to this work.}, David Palzer\footnotemark[1], Jialin Li, Eric Fosler-Lussier, Ningchuan Xiao\\
  The Ohio State Unviersity\\
  \texttt{\{chang.1692, palzer.1, li.7957, fosler-lussier.1, xiao.37\}@osu.edu} \\
}

\begin{document}

\maketitle

\begin{abstract}
Choropleth maps are a common visual representation for region-specific tabular data and are used in a number of different venues (newspapers, articles, etc). These maps are human-readable but are often challenging to deal with when trying to extract data for screen readers, analyses, or other related tasks. Recent research into Visual-Question Answering (VQA) has studied question answering on human-generated charts (ChartQA), such as bar, line, and pie charts. However, little work has paid attention to understanding maps; general VQA models, and ChartQA models, suffer when asked to perform this task. To facilitate and encourage research in this area, we present \textbf{MapQA}\footnote{https://github.com/OSU-slatelab/MapQA}, a large-scale dataset of \textasciitilde 800K question-answer pairs over \textasciitilde 60K map images. Our task tests various levels of map understanding, from surface questions about map styles to complex questions that require reasoning on the underlying data. We present the unique challenges of MapQA that frustrate most strong baseline algorithms designed for ChartQA and general VQA tasks. We also present a novel algorithm, Visual Multi-Output Data Extraction based QA (\textbf{V-MODEQA}) for MapQA. V-MODEQA extracts the underlying structured data from a map image with a multi-output model and then performs reasoning on the extracted data. Our experimental results show that V-MODEQA has better overall performance and robustness on MapQA than the state-of-the-art ChartQA and VQA algorithms by capturing the unique properties in map question answering.

\end{abstract}

\section{Introduction}
\label{sec:intro}
Maps, a common and efficient method to represent statistical tabular data over geographic areas, are deeply-rooted in many daily-life applications, such as representing COVID-19 data or election results. 
A widely-used method to represent such data is the choropleth map type \cite{campbell1998map, white2017trends}. A choropleth map contains a title, a legend, and a set of geographic areas that are colored to represent an aggregate summary of a geographic characteristic within each area. Due to their human-friendly nature compared to raw tabular data, maps often appear without the underlying data in most documents. The immediate application of machine understanding of choropleth maps is as an aid in extracting and understanding the relevant information within the map images. Besides its practical applications, a natural language interface to maps is an interesting study for multi-modal learning as it requires understanding of map images, underlying structured data, and natural language requests. In this paper, we study machine understanding of map images via question answering.

\begin{figure*}
\centering
\subcaptionbox{MapQA-U Sample Map\label{fig:map_U}}
{\includegraphics[width=0.32\textwidth]{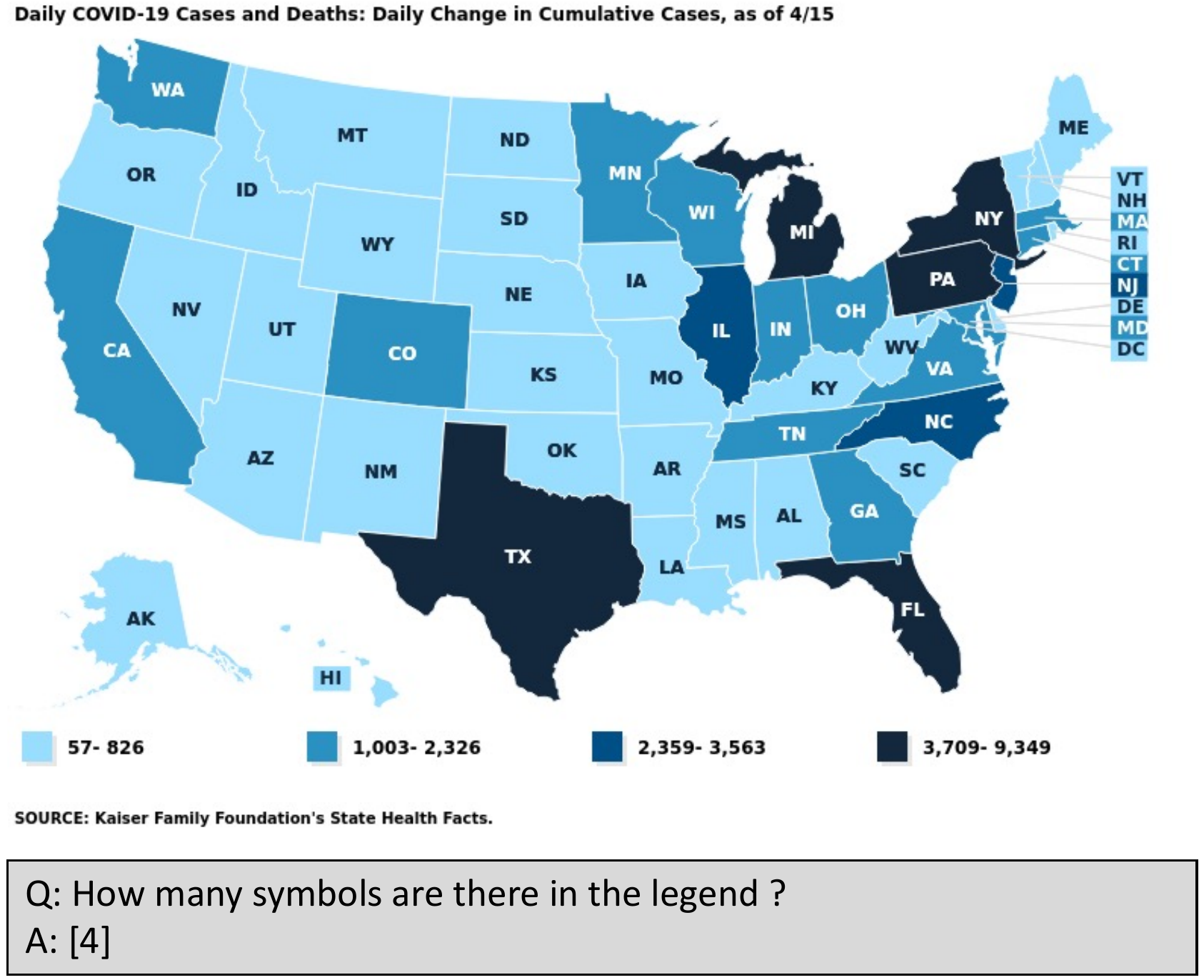}  }
  \subcaptionbox{MapQA-R Sample Map\label{fig:map_R}}  {\includegraphics[width=0.32\textwidth]{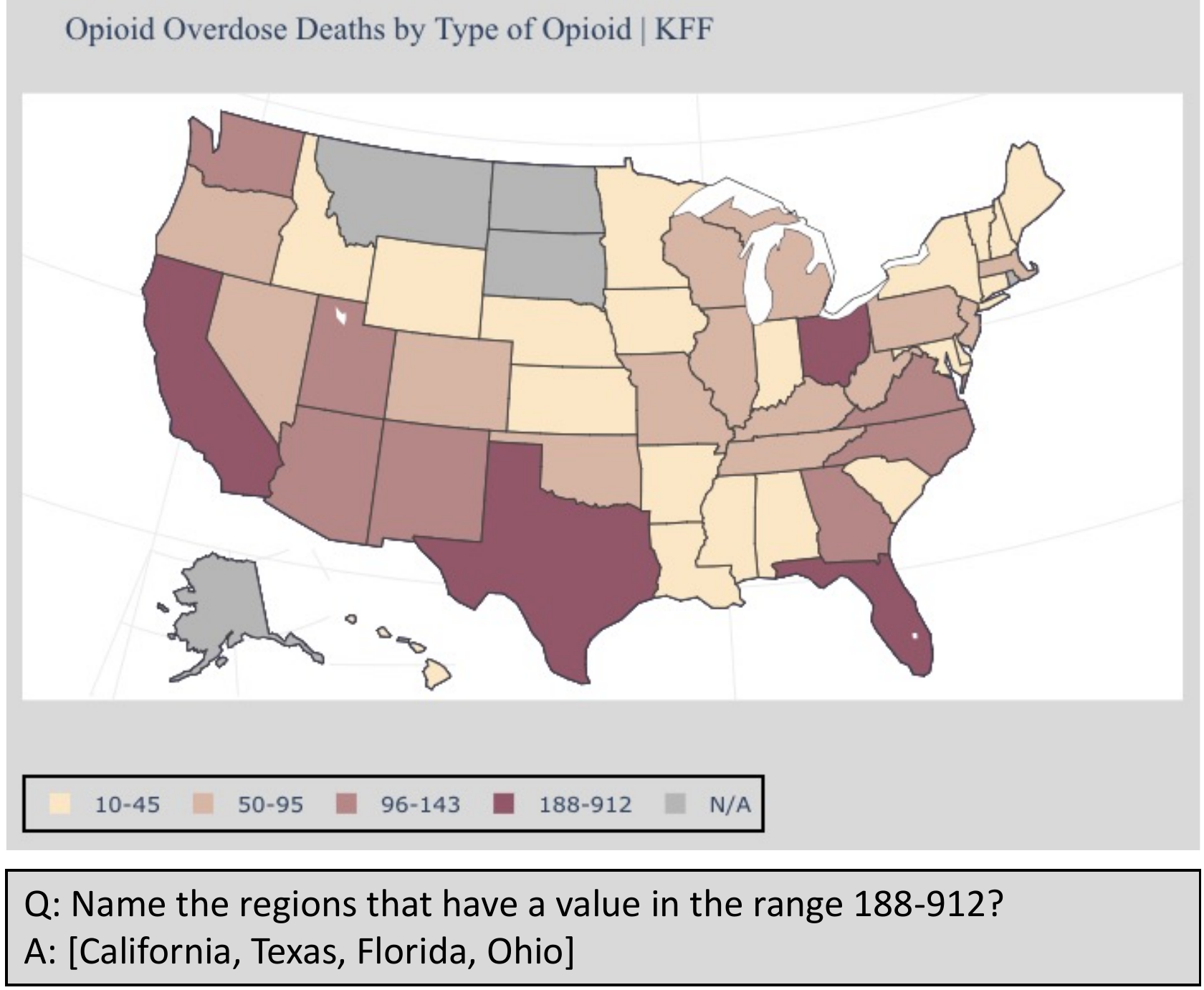}  }
  \subcaptionbox{MapQA-S Sample Map\label{fig:map_S}}
{\includegraphics[width=0.32\textwidth]{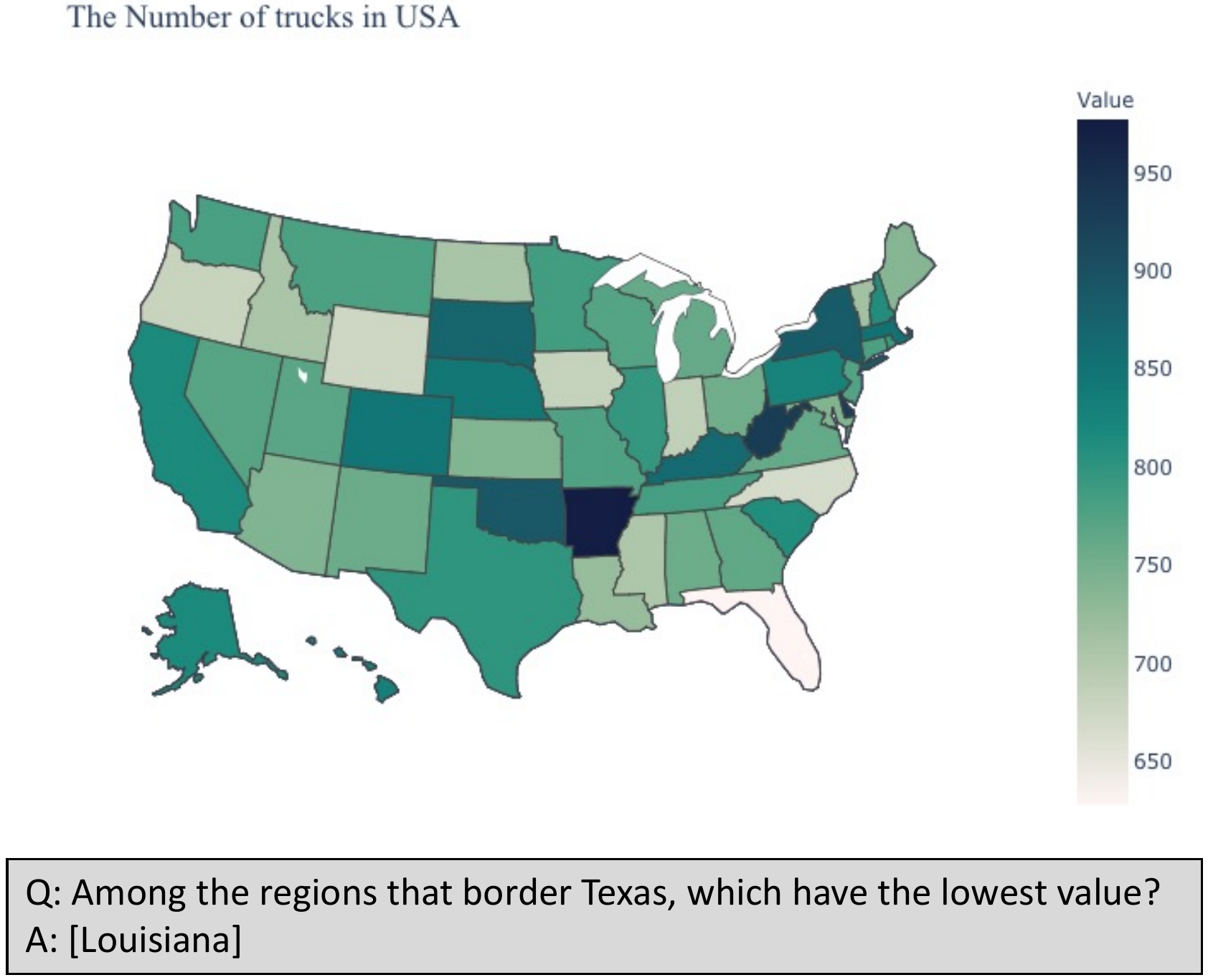}  }

    \caption{ Examples sampled from the MapQA dataset. (a) a real-world map from the Kaiser Family Foundation. (b) a synthetic map generated from the underlying data in KFF maps. (c) a map generated from synthetic data. More examples are included in the supplementary material.}
    \label{fig:examples}
    \vspace{-10pt}
\end{figure*}

Recent years have seen remarkable progress in understanding structured representation, such as relational databases and tables \cite{zhong2017seq2sql, yu2018spider, wang2019rat, scholak2021picard}, however, little work has paid attention to the scenario that structured data are represented in unstructured formats, such as  map images due to the limited number of public datasets. Existing map datasets \cite{gui2016global,hu2021enriching} only contain a couple of thousand map images which only allow machine learning models to learn basic information about maps, such as identifying what area a map represents. 
To address the need for a large dataset, we collect the first large-scale choropleth map QA dataset \textbf{MapQA} which contains around 800K question-answer pairs for over 60K map images. Figure \ref{fig:examples} contains examples in MapQA. The MapQA dataset consists of three subsets: (1) MapQA-U; containing real-world Choropleth maps with a uniform map style, (2) MapQA-R; containing regenerated maps from the underlying data in MapQA-U maps with various different styles, and (3) MapQA-S; containing maps generated from synthetic underlying data with various map styles. MapQA contains various synthetic questions to test the understanding of maps at different abstraction levels, from surface questions related to map styles to complex questions that require reasoning over the underlying data within a map.
To the best of our knowledge, we are the first to create a large-scale choropleth map dataset. MapQA provides a testbed for machine learning models on understanding choropleth maps.

Besides question answering on natural images (VQA), question answering on charts (ChartQA, e.g. bar and pie charts), has gained a relatively large amount of attention recently \cite{kafle2018dvqa,kahou2017figureqa,chaudhry2020leaf, singh2020stl, methani2020plotqa}.
We experiment with strong VQA and ChartQA baselines on the MapQA task. Most of these models regard map question answering as a reasoning task over image and text. However, we believe that understanding maps requires multi-step reasoning: shallow visual reasoning and complex logical reasoning. Visual reasoning represents understanding the map content, including relating each region to its corresponding symbol in the legend and extracting data, while logical reasoning refers to the logical operations over the underlying data such as relational operations and data aggregation.
We find that explicitly representing the tabular data underlying a map image and separating logical reasoning from visual reasoning is beneficial to answering map-related questions. Therefore, we propose a two-stage framework V-MODEQA that extracts the underlying data with a novel visual multi-output data extraction model (V-MODE) and answers questions using the extracted data.
Experimental results demonstrate that the proposed method better captures the multi-modal property of map question answering versus the VQA and ChartQA baselines.

Our contributions in this paper are three-fold: 

\vspace{-4pt}
\begin{itemize}[leftmargin=*]
    \setlength\itemsep{0em}
    \item We construct a large-scale public dataset, MapQA, containing \textasciitilde 800K question-answer pairs grounded in over 60K map images. The three subsets of MapQA can be used to test different aspects of models due to the different representations of the underlying data.
    \item We present the unique challenges of MapQA, compared to VQA and ChartQA, which frustrate strong VQA and ChartQA algorithms. 
    \item We propose a two-stage QA model with a novel multi-output data extraction model that leverages the multi-modal nature of MapQA. Experimental results show it outperforms strong VQA and ChartQA models on map question answering.
\end{itemize}

\vspace{-5pt}
\section{Related Work}
\vspace{-4pt}

\noindent\textbf{Visual Question Answering}
Question answering for natural images (VQA) \cite{antol2015vqa,goyal2017making,krishna2017visual}, as an important task for the research of visual reasoning and multi-modal learning, has been extensively studied \cite{ wu2017visual, xu2015show,yang2016stacked, malinowski2014multi, krishna2017visual}. 
Recently, machine understanding of human-generated data visualizations has generated much interest. Multiple datasets have been publicly released, such as chart QA \cite{kafle2018dvqa,kahou2017figureqa,chaudhry2020leaf,methani2020plotqa}, icon QA \cite{lu2021iconqa}, diagram QA \cite{kembhavi2016diagram, kembhavi2017you} and QA tasks for spatial reasoning \cite{johnson2017clevr,suhr2017corpus}.
From the perspective of input type, ChartQA is the closest task to MapQA; both contain images that are generated from underlying structured data. However, MapQA focuses on maps that have different challenges from other chart images. For example, a map usually contains a larger number of objects (geographic regions) than pie and bar charts, and objects in maps could have complex shapes compared to the simple geometric objects in other charts.

\noindent\textbf{Chart Data Extraction}
ChartQA can also be regarded as two cascaded stages: (1) underlying table extraction and (2) table QA. Previous methods for chart data extraction rely on either predefined geometrical features \cite{poco2017reverse,savva2011revision,gao2012view} or bounding box labels of regions \cite{mishra2021evaginating, kim2020answering,luo2021hybrid,methani2020plotqa}. For example, a bar can be extracted by a rectangle bounding box detection and the height of the rectangle can represent its value. However, such methods are not compatible with MapQA due to the complex object shapes and the lack of bounding box labels. Our proposed multi-output data extraction model does not require such geometrical properties and has the ability to extract continuous and discrete data in parallel.

\noindent\textbf{Database Question Answering}
Questions regarding a choropleth map are usually related to its underlying data. Besides using a map image, another way to represent statistical data of geographic areas is a relational database. However, it is not as efficient as a map for human understanding of the statistical data. Most articles in the wild only release map images to represent their data rather than the raw underlying database. Therefore, understanding a map image cannot be achieved by a database QA system. GEOQuery \cite{zelle1996learning} is a database QA dataset where the database contains geographic information and statistical data in each geographic region. Regarding questions, GEOQuery is the most similar dataset to MapQA, however, it is built for question answering on a database instead of images. Extensive studies \cite{zhong2017seq2sql,yu2018spider,wang2019rat,herzig2020tapas} have done on relational database QA which is able to serve as the logical reasoning module for MapQA.

\noindent\textbf{Map Generation}
There is a scarcity of large-scale datasets for machine understanding of map images due to the limit of copyrights and expensive annotation. Hu et al. \cite{hu2021enriching} use multiple image processing methods, such as projection, scaling, and resizing to generate additional maps. However, such methods can only provide a few thousand map images which, due to the limited size, cannot support large machine learning models in learning a reasonably complex reasoning task. Instead, we synthetically generate map images from both real-world statistical data and synthetic sampled data with common distributions (normal and uniform).

\vspace{-4pt}
\section{Dataset}
\vspace{-4pt}

MapQA tests multiple aspects of understanding a map including surface questions about images, such as understanding the legend type and the number of symbols, to complex questions which require reasoning over the underlying data. The corpus contains 795,525 questions over 62,367 map images in total while it consists of 3 subsets acquired via different approaches as listed below. Table \ref{tab:data_statisitc} contains the data statistics. In each subset, we separate images into defined 60\%/20\%/20\% train/valid/test splits. 

\textbf{MapQA-U} stands for MapQA with a \textbf{U}niform map style. We scrape state-level US map images from Kaiser Family Foundation (KFF) \footnote{https://www.kff.org/statedata/} along with the underlying data. MapQA-U contains 20789 map images from 14 domains and nearly 300K questions. All maps are generated by KFF with a uniform map style. MapQA-U only contains discrete legend maps with the same color scale (the blues in Figure \ref{fig:map_U}), legend position, and orientation. By default, the underlying data contains correlations due to its real-world nature.

\textbf{MapQA-R} represents MapQA with \textbf{R}egenerated KFF maps. MapQA-R contains the same underlying data as MapQA-U while we generate map images from the underlying data instead of using the uniform style maps in MapQA-U to evaluate models on various map styles. To prevent models from overfitting the color scale in interpreting a map (and ignoring the legend), we use disjoint color scales for train/val/test splits in MapQA-R.

\textbf{MapQA-S} denotes MapQA with \textbf{S}ynthetic data. MapQA-S uses the same methods to generate map images as MapQA-R while it uses synthetically generated data over US states. The synthetic underlying data permits minimal answer bias through precise control of the underlying data distribution which makes it harder for models to exploit answer bias while ignoring map content.

\begin{table*}[!htb]
    \centering
    \scriptsize
    \begin{tabular}{lcccccc}
    \toprule
    & \bf DVQA & \bf FigureQA & \bf MapQA & \bf MapQA-U & \bf MapQA-R  & \bf MapQA-S \\
    \midrule
    image type & bar  & bar, line, pie & map & map & map & map\\
    \midrule
    \# images & 300,000  & 180,000 &   62,367 & 20,789  & 20,789 &   20,789 \\
    \midrule
    Disjoint color &   & \checkmark & \checkmark & & \checkmark & \checkmark\\
    \midrule
    Require OCR & \checkmark &  & \checkmark & \checkmark & \checkmark & \checkmark \\
    \midrule
    Max \# objects & 25  & 10 & 50 & 50 & 50 & 50 \\
    \midrule
    \# questions & 3,487,194  & 2,388,698 & 795,525 &  296,084 & 251,018 & 248,423 \\
    \midrule
    \# surface questions & - & - & 86,507 & 31,309 & 27,574 & 27,624 \\
    \midrule
    \# retrieval questions &- & - &313,637 & 124,447 & 95,122 & 94,068 \\
    \midrule
    \# relational questions &- &- &395,381 & 140,328 & 128,322 & 126,731 \\
    \midrule
    percent of yes/no questions  & 23.5\% & 100\% & 27.9\%  & 25.8\% &29.1\% & 29.1\% \\
    \midrule
    multiple answer questions &  &  & \checkmark & \checkmark & \checkmark & \checkmark \\
    \midrule
     \# answers / questions & 1 & 1 & 4.0 &4.0&3.7&4.2 \\
    \midrule
    \# unique answers & 1,576  & 2  & 70,267  & 37,513  & 29,790  & 34,906 \\
    \bottomrule
    \end{tabular}
    \caption{Data statistic of MapQA and comparisons with ChartQA datasets.}
    \label{tab:data_statisitc}
    \vspace{-15pt}
\end{table*}

\subsection{Map Styles and Underlying Data}
\vspace{-4pt}
\label{sec:data_and_map}

We use python's map-drawing tools, Geopandas\footnote{ https://geopandas.org/en/stable/} and Plotly\footnote{https://plotly.com/python} to generate synthetic maps for MapQA-R and MapQA-S. When generating a map we randomly choose one of 55 color scales,\footnote{https://plotly.com/python/builtin-colorscales/} including inverted scales, to represent the underlying data. We consider two strategies for mapping values to colors in cartography: classified maps (with discrete legends) which separate the range of values into classes and all regions in each class are assigned the same color and unclassed maps (with continuous legends) which directly assign a color proportional to the value of each region. For instance, Figure \ref{fig:map_R} is a classified map with a discrete legend, and Figure \ref{fig:map_S} is an unclassed map with a continuous legend. Some other variations include location, orientation, and order of legend; location and font of title; presence or absence of grid lines; and background color.

To generate underlying data in MapQA-S, we consider two types of data: absolute and relative, which are both generated following either uniform or normal distribution in the range of 1-1e8 for absolute data and 1-100 for relative data.
Following the distribution in MapQA-U and MapQA-R, 28\% of maps contain missing values for at least one (or more) region(s).
The colors of regions are continuous for unclassed maps. For classified maps, we first split regions into 2-5 groups by their value based on equal interval or quantile, and then create a description to represent the range in each group. Regions that fall within the same group have the same color. We generate the title with the template ``The Number/Percentage of <NOUNS> in the USA". We randomly choose the word <NOUNS> from the top 500 frequent words in the Brown Corpus tagged Nouns via NLTK \footnote{https://www.nltk.org/}.

\vspace{-3pt}
\begin{table*}[!thb]
    \centering
    \scriptsize
    \begin{tabular}{lc}
    \toprule
    \bf Question Categories & \bf Example Templates  \\
    \midrule
    \multirow{4}{*}{Surface} &  \multicolumn{1}{l}{Is the legend a continuous bar?} \\\cmidrule{2-2}
    &  \multicolumn{1}{l}{How many symbols are there in the legend?} \\\cmidrule{2-2}
    &  \multicolumn{1}{l}{Does the first symbol in the legend represent the smallest category?} \\\cmidrule{2-2}
    &  \multicolumn{1}{l}{Does the map have missing data?} \\\midrule
    \multirow{2}{*}{Retrieval} &  \multicolumn{1}{l}{What is the value of region \textbf{R}?} \\\cmidrule{2-2}
    &  \multicolumn{1}{l}{Name the regions that have a value in the range \textbf{G}} \\\midrule

    \multirow{4}{*}{Relational} &  \multicolumn{1}{l}{Which regions have the lowest / highest value on the map?} \\\cmidrule{2-2}
    &  \multicolumn{1}{l}{What is the lowest / highest value in subarea \textbf{S}?} \\\cmidrule{2-2}
    &  \multicolumn{1}{l}{Among the regions that border \textbf{R\textsubscript{1}}, does region  \textbf{R\textsubscript{2}} have the lowest / highest value?} \\\cmidrule{2-2}
    &  \multicolumn{1}{l}{Does region \textbf{R\textsubscript{1}} have a higher / lower value than region \textbf{R\textsubscript{2}}?} \\
    \bottomrule
    \end{tabular}
    \caption{Question categories and template examples in each category. Note that region refers to a geographic unit, and subarea refers to a collection of regions.}
    \label{tab:question_template}
    \vspace{-10pt}
\end{table*}

\vspace{-4pt}
\subsection{Questions and Answers Generation}
\vspace{-4pt}
\label{sec:question}
We generate questions and their answers by referring to the underlying data. MapQA contains three categories of questions: (1) \textbf{surface questions} which test whether a system can understand overall visual information within maps, the first step of understanding map content; (2) \textbf{retrieval questions} which require understanding maps by extracting the value of individual regions in the maps; and (3) \textbf{relational questions} which require a system to retrieve the values of multiple regions and perform operations on them, adding abstraction relative to retrieval questions. Each category represents a level of reading and understanding maps. Table \ref{tab:question_template} contains template examples in each category. We randomly generate questions from predefined templates. Questions generated from the same template will vary in regard to relevant region, data, and map type.

\noindent \textbf{Post-processing Questions}
For unclassed maps (maps with a continuous legend), 
distinguishing two very close colors is not easy even for humans, also regions are highly unlikely to have the exact same value. Therefore, we do not consider questions that ask whether two regions have the same value.
Additionally, we do not ask models to retrieve exact numbers in continuous legend maps. Those questions are removed during post-processing. For yes/no questions, we balance the ratio of yes and no answers in relational questions. This makes it more difficult for models to exploit answer bias instead of understanding maps.

\subsection{Data Statistics and Evaluation Metric}
\vspace{-4pt}
\label{sec:data_stat}

Table \ref{tab:data_statisitc} contains the data statistics of the three subsets in MapQA. All three subsets have the same number of map images (20,789). The number of questions is slightly different due to the question post-processing within the different sets. We compare our MapQA dataset with two popular ChartQA data sets (DVQA \cite{kafle2018dvqa} and FigureQA \cite{kahou2017figureqa}). To summarize, MapQA contains significantly more unique answers than DVQA and Figure QA. MapQA also requires reasoning across a larger number of objects than ChartQA. Furthermore, MapQA contains multiple-answer questions while the two chartQA data sets do not. For example, a yes/no question has a single answer while a question such as "Which regions have the lowest value in the map" may have multiple answers. We use Jaccard Index to evaluate predicted answers.

\vspace{-4pt}
\section{Methods}
\vspace{-4pt}

We present several baseline models in Section \ref{sec:baseline} followed by the proposed method in Section \ref{sec:MODEQA}. As maps contain important information as text within the image (i.e legend descriptions), optical character recognition (OCR) is required for interpreting a map chart. Following previous chart QA work \cite{kafle2020answering,singh2020stl} we use an OCR model to recognize text within images, which is separate from the QA model. Thus, the QA model can be regarded as solving a classification task. We describe OCR integration for answering map questions in Section \ref{sec:OCR}. Last, Section \ref{sec:training} contains training details.

\begin{figure*}
    \centering    \includegraphics[width=0.95\textwidth]{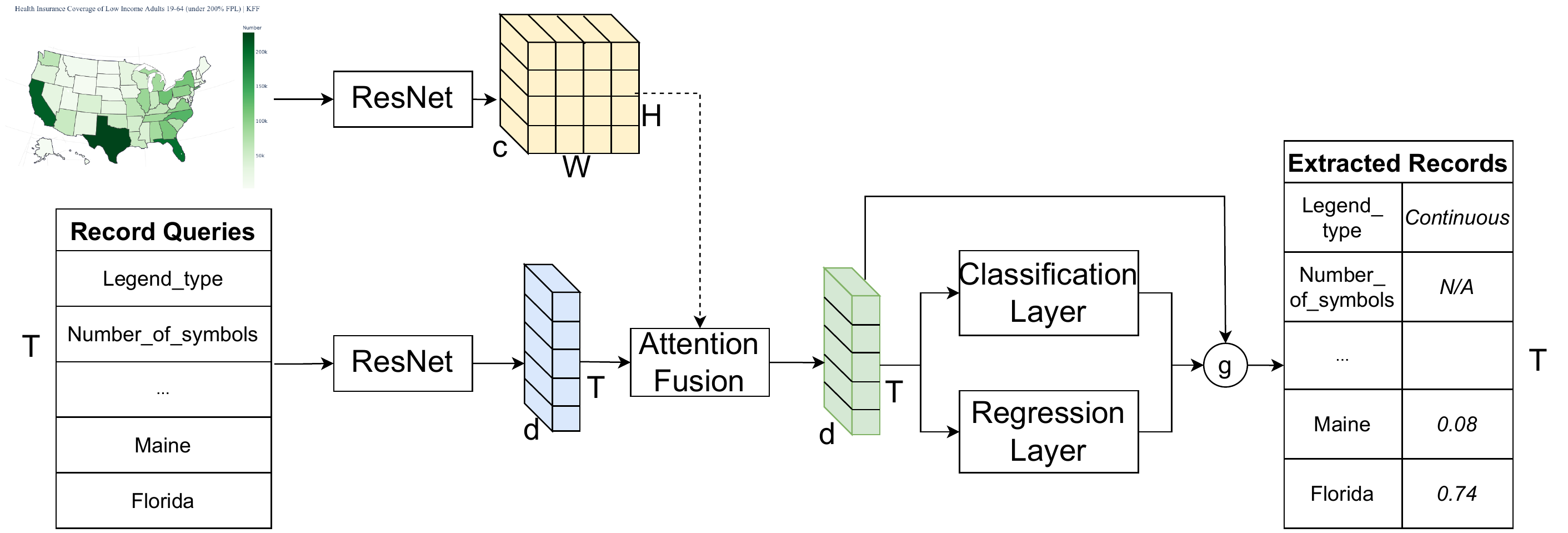}
    \caption{Architecture of our proposed Visual Multi-Output Data Extraction (V-MODE) model. The solid line represents the query of the attention mechanism and the dashed line represents the keys and values.}
    \label{fig:MODE}
    \vspace{-13pt}
\end{figure*}

\vspace{-3pt}
\subsection{Baseline Models}
\vspace{-3pt}
\label{sec:baseline}

\textbf{Blind Methods} are not aware of questions, images, or neither to make the prediction. 
\vspace{-2pt}
\begin{itemize}[leftmargin=*]
    \setlength\itemsep{-0em}
    \item Random: We randomly sample an answer from the answer set in the train split which does not rely on specific text in legends (e.g ``Yes'', ``California'' are legend unspecific).
    \item Majority: This model always predicts the most frequent answer in the train split.
    \item Image: This model is question blind. Images are encoded with a Resnet-152 \cite{he2016deep} and the representation after pooling is used as the final feature set. Two fully-connected (FC) layers are used as the classification model.
    \item Question: This model is image blind. Questions are encoded via a Bi-directional LSTM \cite{hochreiter1997long}. The concatenation of the last hidden vector from both directions is used as the final feature set. A two-layer FC network is used as the classification model.
\end{itemize}
\vspace{-2pt}

\textbf{VQA/ChartQA Models} include strong baselines that are proposed for VQA and ChartQA. These models regard a map as an image without explicitly considering the underlying data. We do not include VQA or ChartQA models that are trained with object bounding boxes \cite{singh2020stl, anderson2018bottom} due to the fact that bounding box labels are not available in MapQA.
\vspace{-2pt}
\begin{itemize}[leftmargin=*]
    \setlength\itemsep{0em}
    \item I+Q Concat: This model combines the model Image and Question from above. It concatenates the question representation and the image representation to form the final feature set. The classification layer is again a two-layer FC network.
    \item SAAA \cite{kazemi2017show}: This model is a previous state-of-the-art (SOTA) model on natural image QA dataset VQA 1.0 \cite{antol2015vqa} and VQA 2.0 \cite{goyal2017making}. SAAA encodes the image and question with a Resnet and LSTM independently and uses a multi-head attention fusion for reasoning over the question and image representation. 
    \item PReFIL \cite{kafle2020answering}: This model is the SOTA model without object bounding box supervision on two large ChartQA datasets: DVQA\cite{kafle2018dvqa} and FigureQA\cite{kahou2017figureqa}. PReFIL extracts the question features from an LSTM model and extracts the image features from both low and high-level layers in a Densenet \cite{huang2017densely}. PReFIL fuses question/image representations via recurrent networks.
\end{itemize}
\vspace{-2pt}

\vspace{-4pt}
\subsection{V-MODEQA}
\vspace{-4pt}

\label{sec:MODEQA}
Reasoning in MapQA contains multiple steps. We consider them in two phrases: shallow visual reasoning and complex logical reasoning. Visual reasoning refers to understanding the map content including connecting each region to a legend symbol and extracting data, and logical reasoning represents logical operations over the underlying data such as relational operations and data aggregation.
Thus, we propose a method V-MODEQA for map question answering that comprises two cascaded stages: (1) Visual MultiOutput Data Extraction and (2) Table QA. We first extract the underlying data from map images with a novel data extraction model then we execute a table QA model to answer questions based on the extracted data.

\vspace{-10pt}

\paragraph{\textbf{Visual Multi-Output Data Extraction}}
Previous studies \cite{ kim2020answering, methani2020plotqa} have used two-stage models for ChartQA, which decouple image-based question answering into data extraction and table-based question answering. Most chart data extraction work \cite{poco2017reverse, kim2020answering,luo2021hybrid} are based on fully utilizing the geometry of objects. For example, a bar object can be detected using rectangular bounding box detection, where the height of the rectangle represents its value. However, such geometric knowledge does not exist in a map image. 
In addition, the underlying data of a map has various types. Surface information of a map can be represented using discrete value, including the legend type, number of legend symbols, legend order, and the appearance of missing data in the map. The value of each region is discrete in classified maps and continuous in unclassed maps. Naturally, we want to have a regression model for continuous value extraction and a classification model for discrete value extraction. 
To extract the various types of values from map images, we propose a novel data extraction model V-MODE standing for \textbf{V}isual \textbf{M}ulti-\textbf{O}utput \textbf{D}ata \textbf{E}xtraction, which is capable of predicting multiple output types, which are discrete categories and continuous numbers. For region value extraction in unclassed maps, we predict the relative position (in range of 0 to 1) of their region colors in the continuous bars. For surface data extraction and classified map region value extraction, we predict their categories.

We formalize the data extraction task and our proposed V-MODE model. Given a map image $I$, and a sequence of tokens $X=\{x_1,...,x_T\}$ which represents a list of record queries, the goal of V-MODE is to extract the corresponding values of each record. For simplification, we use a single token to represent a record query. We have $x_1$ represent the legend type which will be used to combine multiple output types. Following ChartQA work \cite{chaudhry2020leaf,singh2020stl}, we assume underlying data is available for synthetically generated data during training. We refer to $Y^{ext}=\{y^{ext}_1,...,y^{ext}_T\}$ as the gold extraction value for each record. Figure \ref{fig:MODE} shows the architecture of the V-MODE model. We take the output of the ResNet prior to the final pooling layer as our image features $\mathcal{H}^{img} \in \mathbbm{R}^{L\times C}$, where $C$ is the depth dimension of the image features. $L = W \times H$ is the number of spatial dimensions where $W$ is width and $H$ is height. The representation of the extraction sequence $\mathcal{H}^{ext}=\{h^{ext}_1,..,h^{ext}_T\} \in \mathbbm{R}^{N\times D}$ is obtained from the Bi-LSTM model, where $T$ is the length of the extraction sequence and $D$ is the dimension of the hidden layer of Bi-LSTM. We fuse the image and the text features to obtain the combined representation $\mathcal{H}^{comb}=\{h^{comb}_1,...,h^{comb}_T\}$ using a standard attention  mechanism \cite{luong2015effective}.
The combined representation $\mathcal{H}^{comb}$ is fed into a classification output layer and a regression output layer. A binary hard gate $g_i$ is used to combine the predictions from the classification and regression layers. $g_i$ represents the output type of each record $x_i$, which is predicted from the record input and the predicted legend type $\Tilde{y}^{ext}_1$.

\vspace{-4pt}
\begin{align}
P( y_i^{cls} |X,I ) &= Classifier(h_i^{comb}) \\
\Tilde{y}_i^{cls} & =  \arg\max\limits_{k} P( y_i^{cls}=k |X,I ) \\
\Tilde{y}_i^{rgs} &= Regressor(h_i^{comb}) \\
%\Tilde{y_i} &= g_i \cdot \Tilde{y_i} ^{cls} + (1-g_i) \cdot \Tilde{y_i}^{rgs} \\
\Tilde{y}^{ext}_i &= \begin{cases} 
\Tilde{y}_i^{rgs} & g_i=1 \\
      \Tilde{y}_i^{cls} & g_i=0
\end{cases}
\end{align}

We use root mean squared error as the regressor loss $ \mathcal{L}^{rgs}$ and cross entropy as the classifier loss $ \mathcal{L}^{cls}$. The final loss is a weighted summation $\mathcal{L^{\textsc{V-MODE}}} = \mathcal{L}^{rgs} + \lambda \cdot \mathcal{L}^{cls}$.

\paragraph{\textbf{Table QA}}

Our proposed method V-MODEQA is compatible with any Table QA models after extracting the underlying table with V-MODE. For simplification, we use a simple and effective model to answer questions based on the extracted table.
Following previous work on encoding tables \cite{herzig2020tapas, chen2019tabfact}, we flatten the extracted underlying table into a text sequence with a template. For instance, the extracted table in Figure \ref{fig:MODE} is mapped into a sequence ``Legend type is continuous. The number of symbols is N/A. ... Maine's value is 0.08. ...''. We round all real numbers to the hundredths place in the flattened table sequence.

We formally define the Table QA task as follows. Given a question $Q=\{w_1,...,w_N\}$ containing N words and a flattened table sequence $S=\{s_1...,s_P\}$, where $S$ represents T records in P tokens, the goal of Table QA is to predict a set of answers, $Y$, corresponding to the question, $Q$.
After mapping tokens in the question and flattened table sequences to embeddings, two Bi-LSTM models are used to encode the question sequence and flattened table sequence to obtain their representation $\mathcal{H}^{q}=\{h^q_1,...,h^q_N\}$ and $\mathcal{H}^{tab}=\{h^{tab}_1,...,h^{tab}_P\}$. We use an attention mechanism to fuse question and table representation. This along with the average of the question tokens' representations are used as the final features.
%\vspace{-1pt}
\begin{align}
    h^{fuse}_i &= \texttt{Attention}(h^q_i,\mathcal{H}^{tab}) \\
    P(Y|Q,S) &= \texttt{Classifier} ( \underset{i}{\texttt{avg}}(h^{fuse}_i) ) \\ 
    \Tilde{Y} &= \{y|P(Y=y|Q,S)> \xi \}
\end{align}
\noindent where $\Tilde{Y}$ is the predicted answer distribution. Due to the multilabel nature of MapQA task, we use binary cross entropy ($\texttt{BCE}$) as our loss function.
%\vspace{-2pt}
\begin{align}
    \mathcal{L^{\textsc{TableQA}}} &= \texttt{BCE}( P(\Tilde{Y}|Q,S),Y)
\end{align}
\vspace{-3pt}

\vspace{-3pt}    
\subsection{Integrating OCR in MapQA }
\vspace{-4pt}
\label{sec:OCR}
Recognizing the text within a map is an important step in map interpretation. This requires OCR to recognize the text tokens of labels, the legend, or title. Tokens in either the question or answer are from a fixed vocabulary (e.g. "Yes", "Virginia") or are map-specific words (e.g. 3,709-9,349 in Figure \ref{fig:map_U}). 
As we assume the underlying data is available during training, we train all models with an oracle OCR. In testing, we experiment with both the oracle OCR and a real OCR system, Tesseract v5 \footnote{https://github.com/tesseract-ocr/tesseract} \cite{kay2007tesseract}. 
We use the OCR's output to pre-process questions with a heuristic string-matching method. Each symbol and its description in a legend are assigned a position (left-right, top-bottom). Map-specific words in the question are replaced with their position in the legend. For example, 3,709-9,349 is the 4th symbol in Figure \ref{fig:map_U}. The question ``Which states have a number in the range 3,709 - 9,349 ?'' will be pre-processed to ``Which states have a number in the range legend\_4 ?'' Likewise, if the predicted answer is legend\_j we will map the symbol to its description based on OCR results.
For the real OCR system, we obtain all detected strings from Tesseract and separate the legend words from other noisy strings based on their position in the image. A list of detected description phrases is regarded as legend\_1,...,legend\_M.

\vspace{-4pt}
\subsection{Training Details}
\vspace{-4pt}

\label{sec:training}
All of the models that process images use the same image pre-processing, namely resizing the image to 448x448 by padding and resizing, keeping the aspect ratio. The image is normalized based on the image channel mean and standard deviation in the training set. To encourage models to generalize to unseen color scales, we apply a data augmentation during training for all models by randomly shifting the hue channel by up to 20\%.
Because questions are generated from templates, we do not consider large pre-trained language models, such as BERT \cite{devlin2018bert}, to improve question representation. 
For a fair comparison, all models use the same size for question and image encoders. The question encoder is a 300-unit bidirectional LSTM layer\cite{hochreiter1997long} where each word is first embedded into a dense 100-dimensional vector. For models that use a ResNet-152 as the image encoder, the ResNet extracts 14x14x2048 features for each map image. A dropout of 0.3 is used on all FC final layers. 

All question answering models are trained with binary cross entropy as the loss function. Our proposed data extraction model (V-MODE) is trained with both cross-entropy loss and root mean squared error loss. The loss weight $\lambda$ in V-MODE is set to 0.2 to ensure cross-entropy and root mean squared error values exist on the same scale.
All models have been trained for 50 epochs and an additional number of epochs with an early-stopping patience of 20 epochs.
We train all models using the Adam \cite{kingma2014adam} optimizer with a learning rate found using a learning rate search as described in \cite{learningrate}. We also select the prediction threshold based on a grid search on the validation set.

\vspace{-4pt}
\section{Results}
\label{sec:results}
\vspace{-4pt}

\begin{table*}[t]
\centering
%\footnotesize
\scriptsize
%\vspace{-6pt}
\label{tbl:ques-type}
\begin{tabular}{@{}lrrrr|rrrr|rrrr@{} }
\toprule
& \multicolumn{4}{c|}{\textbf{MapQA-U}}       & \multicolumn{4}{c}{\textbf{MapQA-R}}  & \multicolumn{4}{c}{\textbf{MapQA-S}}       \\ 
\multicolumn{1}{c}{} & Srf & Rtr & Rlt & \textbf{All} & Srf & Rtr & Rlt & \textbf{All} & Srf & Rtr & Rlt & \textbf{All} \\ \midrule
    Random  & 1.4 & 0.9 & 1.3 & 1.1 & 1.4 & 0.9 & 1.1 & 1.0 & 1.7 & 0.9 & 1.4 & 1.2  \\
    \midrule
    Majority & 42.0 & 0.0 & 19.0 &  13.5 & 52.8 & 0.0 & 20.0 & 16.0 & 52.8 & 0.0 & 21.1 & 16.7  \\
    \midrule
    Image & 25.0 & 5.2 & 16.9 & 12.8 & 39.0 & 0.0 & 19.8 & 14.3  & 26.3 & 5.0 & 17.1 & 13.6 \\
    \midrule
    Question & 80.4 & 11.8 & 27.9 &  26.7 & 57.6 & 11.7 & 29.2 & 25.6 & 57.6 & 13.3 & 27.2 & 25.4  \\
    \midrule
    \midrule
    I+Q concat w/ Tesseract OCR & 99.9 & 48.0 & 66.8 & 62.4 &  95.8 & 55.0 & 68.0 & 66.1 & 95.8 & 31.4 & 50.6 & 48.4 \\
    I+Q concat w/ Oracle OCR & \textit{99.9} & \textit{75.1} & \textit{81.0} & \textit{80.5} & \textit{95.8} & \textit{59.7} & \textit{69.4} & \textit{68.6}  & \textit{95.8} & \textit{33.1} & \textit{51.7} & \textit{49.6}  \\    
    \midrule
    SAAA w/ Tesseract OCR & 99.9 & 61.1 & 77.1 & 72.8  & 97.0 & 59.5 & 71.5 & 69.7 & 96.2 & 30.5 & 50.4 & 48.0  \\
    SAAA w/ Oracle OCR & \textit{99.9} & \textit{97.8} & \textit{92.4} & \textit{95.5} & \textit{97.0} & \textit{64.5} & \textit{73.0} & \textit{72.4} & \textit{96.2} & \textit{31.9} & \textit{51.5} & \textit{49.1} \\
    \midrule
    PReFIL w/ Tesseract OCR & \bf 100.0 & \bf 61.1 & 79.2 & 73.8 &  95.9 & 75.2 & 75.5  & 77.6 & 97.7  & 67.6 & 64.2 & 69.2 \\
    PReFIL w/ Oracle OCR & \bf \textit{100.0}  & \bf \textit{99.9} & \textit{94.5} & \textit{97.4} & \textit{95.9} & \textit{83.0} & \textit{76.9} & \textit{81.3} &  \textit{97.6} & \textit{73.7} & \textit{65.4} & \textit{72.2} \\
    \midrule
    V-MODEQA w/ Tesseract OCR & \bf 100.0 & 60.7 & \bf 81.0 & \bf 74.5 & \bf 98.7 & \bf 84.6 & \bf 83.9 & \bf 85.8 & \bf 98.0 & \bf 86.3 & \bf 77.9 & \bf 83.3  \\
    V-MODEQA w/ Oracle OCR & \bf \textit{100.0} & \bf \textit{99.9} & \bf \textit{96.8} & \bf \textit{98.4} & \bf \textit{98.7} & \bf \textit{93.1} & \bf \textit{85.7} & \bf \textit{89.9} & \bf \textit{98.0} & \bf \textit{94.5} & \bf \textit{79.2} & \bf \textit{87.1} \\
    \midrule
    \midrule
    Human & 100.0 & 99.8 & 93.5 & 97.8 & 100 & 97.9 & 89.3 & 92.4  & 87.5 & 98.4 & 85.4 & 89.8 \\
    \bottomrule
\end{tabular}
\caption{MapQA Results (Jaccard Index) [\%]. Srf, Rtr, and Rlt represents surface, retrieval, and relational questions. The upper section contains the results of blind baselines which represents the bias of the MapQA dataset. The middle section includes previous VQA and ChartQA state-of-the-art models' results as well as our proposed model. We evaluate them based on a real OCR system, Tesseract, and an oracle OCR. The best results with both Oracle OCR and Tesseract OCR are bold. The bottom section contains manually answered human results. }
\label{tab:results}
\vspace{-15pt}
\end{table*}

%In this section, we present the results of various baseline models and our proposed MODEQA methods on three MapQA subsets. 
This section presents the experiments which aim to answer a series of research questions. Below, we present the questions and attempt to answer them via our conducted experiments.

\textbf{What do we learn about the MapQA dataset from blind models?}
The upper section in Table \ref{tab:results} contains the results of blind baseline methods. Randomly selecting answers from a legend-unspecific vocabulary results in a 1.1\% Jaccard index. Answer ``No'' is the most common answer in MapQA which has a slightly higher frequency than ``Yes''. Predicting the majority answer from the train set obtains less than a 17\% Jaccard index on all subsets, which indicates the bias of the answer set.
The image-only model achieves a similar result to always predicting the majority answer due to the fact it can only capture the answer distribution without knowledge of questions. The question-only model outperforms the Majority model by predicting the most likely answer per question type. It achieves 80.4\% Jaccard index for surface questions in MapQA-U and 57.6\% for surface questions in MapQA-R and MapQA-S. This shows the value of various map styles in MapQA-R and MapQA-S compared to MapQA-U.

\textbf{What challenges do previous state-of-the-art VQA and ChartQA models face on MapQA?}
We find the image and question concatenation model (I+Q concat) performs significantly better than image-only and question-only models, which shows MapQA requires an understanding of both inputs for reasoning. On MapQA-U, concatenation performs poorly for fusing image features and question features compared to the attention mechanism in SAAA and recurrent networks in PReFIL. This indicates the value of MapQA for studying multi-modal learning. Not surprisingly, all models perform better on MapQA-U than on MapQA-R, demonstrating that generalizing to various map styles is challenging. With an Oracle OCR, SAAA obtains 95.5\% and 72.4\% Jaccard Index on MapQA-U and MapQA-R, and PReFIL achieves 97.4\% and 81.3\% Jaccard Index on MapQA-U and MapQA-R. However, they only obtain 49.1\%, 72.2\% on MapQA-S, 23.3\% and 9.1\% lower than MapQA-R. (We report the absolute differences in the rest of the paper.) This indicates that these models shortcut understanding map content by exploiting the underlying data distribution. Real-world statistical data usually contains biases for different geographic regions. For example, as in Figure \ref{fig:map_U} COVID-19 case numbers in New York are more likely to be higher than in Montana due to its larger population. Note that exploiting the underlying data distribution is useful in real applications, however, this can make a model prone to making mistakes when the bias of the underlying data changes. Human understanding of maps usually relies directly on the map content instead of bias within geographic regions. Therefore, we believe MapQA-R and MapQA-S can test models from different perspectives. It is important for models to learn well on both sets.

\textbf{How does the proposed V-MODEQA method perform on MapQA?}
Referring back to Table \ref{tab:results}, our proposed V-MODEQA model achieves 98.4\% on MapQA-U, 89.9\% on MapQA-R, and 87.1\% on MapQA-S with Oracle OCR which are the best results on all three subsets. The results demonstrate the benefit of explicitly modeling the underlying structure data in the map question answering task. Different from baseline models which have significantly lower performance on MapQA-S compared to MapQA-R, V-MODEQA has consistent performance on MapQA-S and MapQA-R. We believe that it is because extracting underlying data explicitly encourages the model to read the map content with its legend instead of capturing the underlying data distribution.

Table \ref{tab:data-extraction-acc} contains the accuracy of V-MODE data extraction. For discrete legend maps, accuracy is the ratio of correct predictions (i.e. the predicted class is the same as the label) to all predictions. For continuous legend maps, labels and predictions are real numbers. We define an evaluation metric $\textsc{Distance}@k$ based on the distance between predicted value $\Tilde{y}^{ext}_i$ and label $y^{ext}_i$: 
\vspace{-5pt}
$$
    \textsc{Distance}@k= \frac{1}{|Y^{ext}|} \sum\limits_{y^{ext}_i \in Y^{ext}} \mathbbm{1}_ {|\Tilde{y}^{ext}_i-y^{ext}_i|<k},
$$
\vspace{-5pt}

\noindent where $\mathbbm{1}$ is the indicator function. $\textsc{Distance}@k$ represents the percentage of predictions that have a distance from the gold value below k. We report k of 0.01, 0.05, and 0.1. V-MODE achieves near 95\% accuracy for discrete legend maps in MapQA-R and MapQA-S. For continuous legend maps, the performance of V-MODE on MapQA-R and MapQA-S is also close, indicating that V-MODE focuses on learning visual content instead of exploiting underlying data bias.
Table \ref{tab:ablation} contains an ablation study on MapQA-S. - multi-output represents removing the multi-output model. Instead of the proposed V-MODE model, we first train a model to predict the legend type, then we separately train one classification and one regression model for continuous and discrete legend maps, respectively. Without our proposed multi-output model, the data extraction accuracy on discrete legend maps drops 5.8\% and \textsc{Distance}@0.1 on continuous legend maps drops 5.7\%, which makes the overall QA result drop 5.1\%. 
We also remove the data augmentation of randomly adjusting the hue channel of images in training. The performance drop illustrates that it encourages models to generalize to unseen color scales.

\begin{table}
    \vspace{-10pt}
    \centering
    \small
    \begin{tabular}{lccc}
    \toprule
    Legend Type & \bf MapQA-U & \bf MapQA-R & \bf MapQA-S   \\
    \midrule
    Continuous  & - & 15.4/58.2/84.0  & 17.3/54.0/91.0 \\
    \midrule
    Discrete  & 99.9 & 92.0 & 94.7 \\
    \bottomrule
    \end{tabular}
    \caption{Data extraction accuracy of V-MODE. For discrete legend maps, we report the classification accuracy. For continuous legend maps, we evaluate the model by the distance between the predicted value and the gold value. x/y/z represents the percentage of data for distance <= 0.01/0.05/0.1.}
    \label{tab:data-extraction-acc}
    \vspace{-2pt}
\end{table}

\begin{table}
\vspace{-5pt}
\small
    \centering
    \begin{tabular}{lcc|c}
    \toprule
    & \multicolumn{2}{c|}{\textbf{Data Extraction Acc}}       & \multicolumn{1}{c}{\textbf{QA Acc}} \\ 
    \bf Model &  Continuous &  Discrete   \\
    \midrule
    V-MODEQA  &   15.3/51.9/89.3 &  94.9 & 87.1 \\
    \midrule
    - multi-output  & 13.9/64.3/83.6 &  89.1 & 82.0\\
    \midrule
    - data augmentation  &7.4/34.1/60.4& 89.8 & 83.3 \\ 
    \bottomrule
    \end{tabular}
    \caption{Ablation study of V-MODEQA on MapQA-S.}
    \label{tab:ablation}
    \vspace{-10pt}
\end{table}

\textbf{How does the OCR system affect question answering?}
MapQA requires understanding the text in the image to answer questions as some important information is expressed as text in the legend. The Jaccard index of V-MODEQA with a real OCR system, Tesseract, is 4.1\% and 3.5\% lower than with Oracle OCR on MapQA-R and MapQA-S, respectively. Tesseract performs poorly on MapQA-U which has a 24.0\% lower Jaccard Index than Oracle OCR. We report the accuracy of recognizing the text of each legend symbol in discrete legend maps in Table \ref{tab:ocr-acc}. The gap between Tesseract OCR and Oracle OCR indicates there is room for improving OCR models for MapQA. OCR in MapQA-U is more challenging than in MapQA-R and MapQA-S due to the relatively lower resolution.

\textbf{How do humans perform on MapQA?}
%\textit{\textbf{Ans}}:
The bottom of table \ref{tab:results} contains a study of human performance on MapQA. Four students with US geographic knowledge answer 50 random questions from each of the three subsets. In MapQA-R and MapQA-S, humans show superior abilities in retrieval and relational reasoning versus machine learning models. Compared to machine understanding of maps, human performance is not limited by OCR accuracy. However, we find that a common mistake of the human baseline is misunderstanding questions. For example, annotators confuse ``What is the highest value'' with ``Which states have the highest value''. Since models do not skim when reading, we believe such mistakes are unlikely to be made.

\begin{table}[!htb]
\small
    \centering
    \begin{tabular}{lccc}
    \toprule
    OCR Model & \bf MapQA-U & \bf MapQA-R & \bf MapQA-S   \\
    \midrule
    Tesseract  & 48.7 & 92.3 & 94.1  \\
    \bottomrule
    \end{tabular}
    \caption{Accuracy of recognizing text of legend symbols for discrete legend maps. }
    \label{tab:ocr-acc}
    \vspace{-2pt}
\end{table}

\vspace{-6pt}
\section{Conclusion and Future Work}
\vspace{-4pt}

This paper introduces MapQA, a large-scale image-question dataset for understanding choropleth maps. MapQA contains three different subsets to evaluate models’ capability of map understanding: 1) real-world data with journalism style maps, 2) real-world data with various styles, and 3) de-biased synthetic data with various styles. Paired with synthetically generated questions, MapQA evaluates models for understanding multiple modalities, including natural language questions, map images, and underlying structured data. We also present a novel model, V-MODEQA, which extracts tabular data underlying map images using a multi-output data extraction model and then executes a table QA model to answer questions based on the extracted data.

In the future, we plan to extend MapQA to additional geographic areas beyond the United States and to other object scales (e.g. counties, cities, etc). We will test the generalization capability of models on real-world choropleth maps by collecting additional real-world maps with varying styles, and crowd sourcing question-answer pairs. We regard MapQA as the first step in exploring machine learning methods for understanding structured data in choropleth maps. 
Our experiments show the benefit of explicitly modeling the underlying structured data. However, the two cascaded stages have cumulative errors. Therefore, we plan to explore learning question answering with data extraction end-to-end in the future.

\clearpage

\bibliographystyle{plainnat}
\bibliography{neurips_2022}

%%%%%%%%%%%%%%%%%%%%%%%%%%%%%%%%%%%%%%%%%%%%%%%%%%%%%%%%%%%%

% \appendix

% \section{Appendix}

% Optionally include extra information (complete proofs, additional experiments and plots) in the appendix.
% This section will often be part of the supplemental material.

\end{document}